\documentclass[11pt,a4paper]{article}
\usepackage[hyperref]{emnlp2018}
\usepackage{times}
\usepackage{latexsym}

\usepackage{mathtools}
\usepackage{graphicx}
\usepackage{amsmath}
\usepackage{adjustbox}
\usepackage{caption}
\usepackage{booktabs}
\usepackage{makecell}
\usepackage{multirow}
\usepackage{amsmath,environ}
\usepackage{subfigure}
\usepackage{amssymb}
\usepackage{pifont}

\usepackage{url}

\aclfinalcopy 

\title{Learning to Encode Text as Human-Readable Summaries using Generative Adversarial Networks} 

\author{Yau-Shian Wang \\
  National Taiwan University\\
  {\tt king6101@gmail.com} \\\And
  Hung-Yi Lee \\
  National Taiwan University \\
  {\tt tlkagkb93901106@gmail.com} \\}

\date{}

\begin{document}
\maketitle
\begin{abstract}
Auto-encoders compress input data into a latent-space representation and reconstruct the original data from the representation. 
This latent representation is not easily interpreted by humans.
In this paper, we propose training an auto-encoder that encodes input text into human-readable sentences, and unpaired abstractive summarization is thereby achieved.
The auto-encoder is composed of a generator and a reconstructor.
The generator encodes the input text into a shorter word sequence, and the reconstructor recovers the generator input from the generator output.
To make the generator output human-readable, a discriminator restricts the output of the generator to resemble human-written sentences.
By taking the generator output as the summary of the input text, abstractive summarization is achieved without document-summary pairs as training data.
Promising results are shown on both English and Chinese corpora.
\end{abstract}

\section{Introduction}\vspace{-2mm}
When it comes to learning data representations, a popular approach involves the auto-encoder architecture, which compresses the data into a latent representation without supervision.
In this paper we focus on learning text representations.
Because text is a sequence of words, to encode a sequence, a sequence-to-sequence (seq2seq) auto-encoder~\citep{hier,skip} is usually used, in which a RNN is used to encode the input sequence into a fixed-length representation, after which another RNN is used to decode the original input sequence given this representation.

Although the latent representation learned by the seq2seq auto-encoder can be used in downstream applications, it is usually not human-readable. A human-readable representation should comply the rule of human grammar and can be comprehended by human. Therefore, in this work, we use comprehensible natural language as a latent representation of the input source text in an auto-encoder architecture. This human-readable latent representation is shorter than the source text; in order to reconstruct the source text, it must reflect the core idea of the source text. Intuitively, the latent representation can be considered a summary of the text, so unpaired abstractive summarization is thereby achieved. 

The idea that using human comprehensible language as a latent representation has been explored on text summarization, but only in a semi-supervised scenario. 
Previous work~\citep{language} uses a prior distribution from a pre-trained language model to constrain the generated sequence to natural language.  
However, to teach the compressor network to generate text summaries, the model is trained using labeled data. 
In contrast, in this work we need no labeled data to learn the representation.

\begin{figure}[ht]
  \centering
    \includegraphics[width=1.0\linewidth]{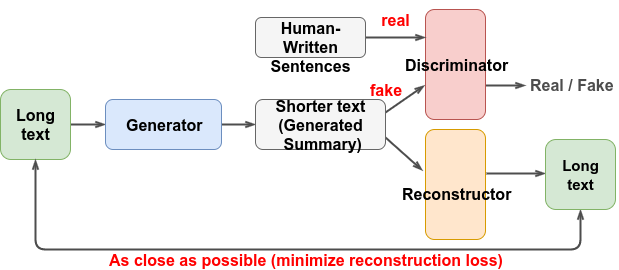}
  \caption{Proposed model. Given long text, the generator produces a shorter text as a summary.
  The generator is learned by minimizing the reconstruction loss together with the reconstructor and making discriminator regard its output as human-written text.}
  \label{fig:big-overview}
\end{figure}

As shown in Fig.~\ref{fig:big-overview}, the proposed model is composed of three components: a generator, a discriminator, and a reconstructor.
Together, the generator and reconstructor form a text auto-encoder.
The generator acts as an encoder in generating the latent representation from the input text. 
Instead of using a vector as latent representation, however, the generator generates a word sequence much shorter than the input text.
From the shorter text, the reconstructor reconstructs the original input of the generator. 
By minimizing the reconstruction loss, the generator learns to generate short text segments that contain the main information in the original input.
We use the seq2seq model in modeling the generator and reconstructor because both have input and output sequences with different lengths.

However, it is very possible that the generator's output word sequence can only be processed and recognized by the reconstructor but is not readable by humans.
Here, instead of regularizing the generator output with a pre-trained language model~\citep{language}, we borrow from adversarial auto-encoders~\citep{AA} and cycle GAN~\citep{cycle} and introduce a third component~-- the discriminator~-- to regularize the generator's output word sequence.
The discriminator and the generator form a generative adversarial network (GAN)~\citep{GAN}.
The discriminator discriminates between the generator output and human-written sentences, and the generator produces output as similar as possible to human-written sentences to confuse the discriminator. 
With the GAN framework, the discriminator teaches the generator how to create human-like summary sentences as a latent representation.
However, due to the non-differential property of discrete distributions, generating discrete distributions by GAN is challenging. To tackle this problem, in this work, we proposed a new kind of method on language generation by GAN.

By achieving unpaired abstractive text summarization, machine is able to unsupervisedly extract the core idea of the documents.
This approach has many potential applications.
For example, the output of the generator can be used for the downstream tasks like document classification and sentiment classification.
In this study, we evaluate the results on an abstractive text summarization task.
The output word sequence of the generator is regarded as the summaries of the input text.
The model is learned from a set of documents without summaries.
As most documents are not paired with summaries, for example the movie reviews or lecture recordings, this technique makes it possible to learn summarizer to generate summaries for these documents.
The results show that the generator generates summaries with reasonable quality on both English and Chinese corpora.

\setlength{\belowcaptionskip}{-20pt}

\section{Related Work}\vspace{-2.0mm}

\addcontentsline{toc}{section}{Unnumbered Section}
\subsection*{Abstractive Text Summarization}\vspace{-2.0mm}
Recent model architectures for abstractive text summarization basically use the sequence-to-sequence~\citep{seq2seq} framework in combination with various novel mechanisms. 
One popular mechanism is attention~\citep{attn}, which has been shown helpful for summarization~\citep{bey,Rush-attn,RAS}.
It is also possible to directly optimize evaluation metrics such as ROUGE~\citep{ROUGE} with reinforcement learning~\citep{RLseq,RLsum,ACseq}.
The hybrid pointer-generator network~\citep{getto} selects words from the original text with a pointer~\citep{pointer} or from the whole vocabulary with a trained weight. 
In order to eliminate repetition, a coverage vector~\citep{coverage} can be used to keep track of attended words, and coverage loss~\citep{getto} can be used to encourage model focus on diverse words.
While most papers focus on supervised learning with novel mechanisms, in this paper, we explore unsupervised training models. \vspace{-2.0mm}

\addcontentsline{toc}{section}{Unnumbered Section}
\subsection*{GAN for Language Generation}\vspace{-2.0mm}
In this paper, we borrow the idea of GAN to make the generator output human-readable. 
The major challenge in applying GAN to sentence generation is the discrete nature of natural language.
To generate a word sequence, the generator usually has non-differential parts such as \textit{argmax} or other sample functions which cause the original GAN to fail.

\vspace{-1.5mm}
In~\citep{improved}, instead of feeding a discrete word sequence, the authors directly feed the generator output layer to the discriminator. 
This method works because they use the earth mover's distance on GAN as proposed in~\citep{wgan}, which is able to evaluate the distance between a discrete and a continuous distribution. 
SeqGAN~\citep{seqgan} tackles the sequence generation problem with reinforcement learning. Here, we refer to this approach as adversarial REINFORCE.
However, the discriminator only measures the quality of whole sequence, and thus the rewards are extremely sparse and the rewards assigned to all the generation steps are all the same. 
MC search~\citep{seqgan} is proposed to evaluate the approximate reward at each time step, but this method suffers from high time complexity.
Following this idea, \citep{advdialogue} proposes partial evaluation approach to evaluate the expected reward at each time step. 
In this paper, we propose the self-critical adversarial REINFORCE algorithm as another way to evaluate the expected reward at each time step. 
The performance between original WGAN and proposed adversarial REINFORCE is compared in experiment. \vspace{-1.0mm}

\section{Proposed Method}\vspace{-1.0mm}

\begin{figure*}[ht]
  \centering
    \includegraphics[width=1.0\linewidth]{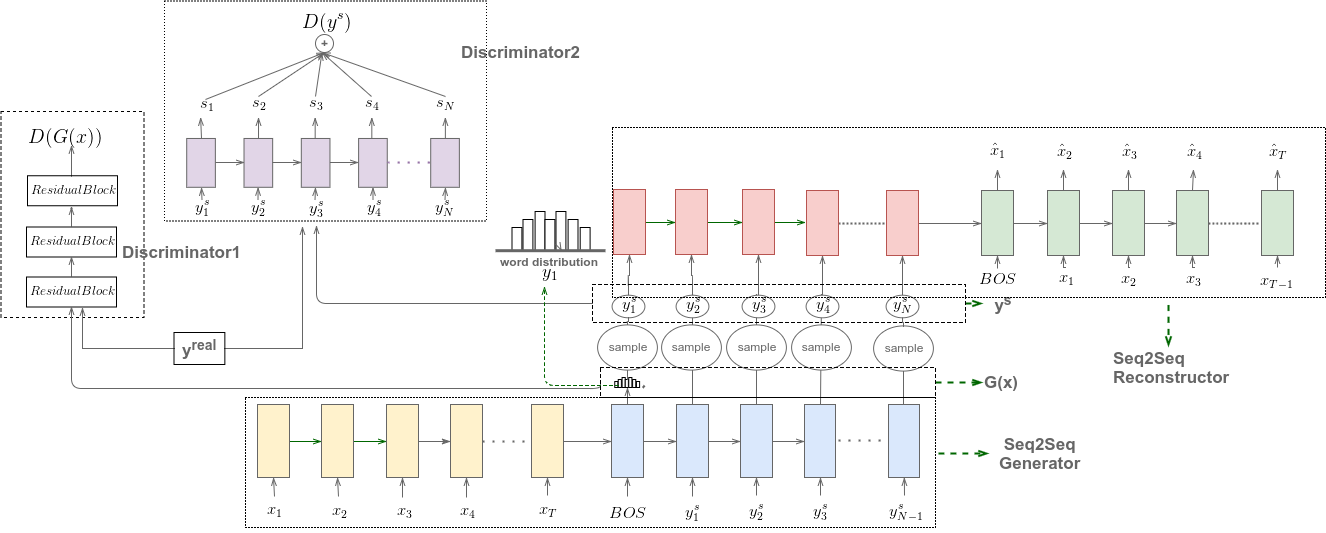}
  \caption{Architecture of proposed model. The generator network and reconstructor network are a seq2seq hybrid pointer-generator network, but for simplicity, we omit the pointer and the attention parts.}
  \label{fig:overview_pic}
\end{figure*}

The overview of the proposed model is shown in Fig.~\ref{fig:overview_pic}.
The model is composed of three components: generator $G$, discriminator $D$, and reconstructor $R$.
Both $G$ and $R$ are seq2seq hybrid pointer-generator networks~\citep{getto} which can decide to copy words from encoder input text via pointing or generate from vocabulary.
They both take a word sequence as input and output a sequence of word distributions. 
Discriminator $D$, on the other hand, takes a sequence as input and outputs a scalar.
The model is learned from a set of documents $x$ and human-written sentences $y^{real}$.
\vspace{-1.0mm} 

To train the model, a training document $x=\{x_{1},x_{2},...,x_{t},...,x_{T}\}$, where $x_t$ represents a word, is fed to $G$, which outputs a sequence of word distributions $G(x)=\{y_{1},y_{2},...,y_{n},...,y_{N}\}$, where $y_{n}$ is a distribution over all words in the lexicon. 
Then we randomly sample a word $y^{s}_{n}$ from each distribution $y_{n}$, and a word sequence $y^{s}=\{y^{s}_{1},y^{s}_{2},...,y^{s}_{N}\}$  is obtained according to $G(x)$. 
We feed the sampled word sequence $y^{s}$ to reconstructor $R$, which outputs another sequence of word distributions $\hat{x}$. 
The reconstructor $R$ reconstructs the original text $x$ from $y^{s}$.
That is, we seek an output of reconstructor $\hat{x}$ that is as close to the original text $x$ as possible; hence the loss for training the reconstructor, $R_{loss}$, is defined as:
\begin{equation}
R_{loss} = \sum_{k=1}^K l_s(x,\hat{x}), \label{eq:Rloss}
\end{equation}
where the reconstruction loss $l_s(x,\hat{x})$ is the cross-entropy loss computed between the reconstructor output sequence $\hat{x}$ and the source text $x$, or the negative conditional log-likelihood of source text $x$ given word sequence $y^{s}$ sampled from $G(x)$.
The reconstructor output sequence $\hat{x}$ is teacher-forced by source text $x$.
The subscript $_{s}$ in  $l_s(x,\hat{x})$ indicates that $\hat{x}$  is reconstructed from $y^{s}$.
$K$ is the number of training documents, and (\ref{eq:Rloss}) is the summation of the cross-entropy loss over all the training documents $x$.\vspace{-1.0mm} 

In the proposed model, the generator $G$ and reconstructor $R$ form an auto-encoder.
However, the reconstructor $R$ does not directly take the generator output distribution $G(x)$ as input
\footnote{We found that if the reconstructor $R$ directly  takes $G(x)$ as input, the generator $G$ learns to put the information about the input text in the distribution of $G(x)$, making it difficult to sample meaningful sentences from $G(x)$.}.
Instead, the reconstructor takes a sampled discrete sequence $y^{s}$ as input.
Due to the non-differentiable property of discrete sequences, we apply the REINFORCE algorithm, which is described in Section~\ref{decoder}.

In addition to reconstruction, we need the discriminator $D$ to discriminate between the real  sequence $y^{real}$ and the generated sequence $y^s$ to regularize the generated sequence satisfying the summary distribution.
$D$ learns to give $y^{real}$ higher scores while giving $y^s$ lower scores.
The loss for training the discriminator $D$ is denoted as $D_{loss}$; this is further described in Section~\ref{GAN}.

$G$ learns to minimize the reconstruction loss $R_{loss}$, while maximizing the loss of the discriminator $D$ by generating a summary sequence $y^s$ that cannot be differentiated by $D$ from the real thing.
The loss for the generator $G_{loss}$ is \vspace{-1.0mm}
\begin{equation}
G_{loss} = \alpha R_{loss} - D_{loss}^\prime \label{eq:Gloss}
\end{equation}
where $D_{loss}^\prime$ is highly related to $D_{loss}$~-- but not necessary the same\footnote{$D_{loss}^\prime$ has different formulations in different approaches. This will be clear in Sections~\ref{IMP} and \ref{ARL}.}~-- and $\alpha$ is a hyper-parameter. 
After obtaining the optimal generator  by minimizing (\ref{eq:Gloss}), we use it to generate summaries.

Generator $G$ and discriminator $D$ together form a GAN.
We use two different adversarial training methods to train $D$ and $G$;  as shown in Fig.~\ref{fig:overview_pic}, these two methods have their own discriminators 1 and 2. 
Discriminator 1 takes the generator output layer $G(x)$ as input, whereas discriminator 2 takes the sampled discrete word sequence $y^{s}$ as input.
The two methods are described respectively in Sections~\ref{IMP} and \ref{ARL}.\vspace{-2.0mm} 

\section{Minimizing Reconstruction Loss}\label{decoder}\vspace{-2.0mm}
Because discrete sequences are non-differentiable, we use the REINFORCE algorithm. 
The generator is seen as an agent whose reward given the source text $x$ is  $-l_s(x,\hat{x})$.
Maximizing the reward is equivalent to minimizing the  reconstruction loss $R_{loss}$ in (\ref{eq:Rloss}).
However, the reconstruction loss varies widely from sample to sample, and thus the rewards to the generator are not stable either. 
Hence we add a baseline to reduce their difference. 
We apply self-critical sequence training \citep{selfcritic}; the modified reward $r^{R}(x,\hat{x})$ from reconstructor $R$ with the baseline for the generator is\vspace{-1.0mm} 
\begin{equation} \label{eq:R_reward}
  r^{R}(x,\hat{x}) = -l_{s}(x,\hat{x}) - (-l_{a}(x,\hat{x}) - b)
\end{equation}
where $-l_{a}(x,\hat{x}) - b$ is the baseline. 
$l_{a}(x,\hat{x})$ is also the same cross-entropy reconstruction loss as $l_{s}(x,\hat{x})$, except that $\hat{x}$ is obtained from $y^{a}$ instead of $y^{s}$.
$y^{a}$ is a word sequence $\{y^{a}_{1},y^{a}_{2},...,y^{a}_{n},...,y^{a}_{N}\}$, where $y^a_n$ is selected using the \textit{argmax} function from the output distribution of generator $y_n$.
As in the early training stage, the sequence $y^{s}$ 
barely yields higher reward than
sequence $y^{a}$, to encourage exploration we introduce the second baseline score $b$, which gradually decreases to zero.
Then, the generator is updated using the REINFORCE algorithm with reward $r^{R}(x,\hat{x})$ to minimize $R_{loss}$.\vspace{-2.0mm} 

\section{GAN Training}\label{GAN}\vspace{-2.0mm} 
With adversarial training, the generator learns to produce sentences as similar to the human-written sentences as possible. Here, we conduct experiments on two kinds of methods of language generation with GAN. In Section~\ref{IMP} we directly feed the generator output probability distributions to the discriminator and use a Wasserstein GAN (WGAN) with a gradient penalty.
In Section~\ref{ARL}, we explore adversarial REINFORCE, which feeds sampled discrete word sequences to the discriminator and evaluates the quality of the sequence from the discriminator for use as a reward signal to the generator.

\subsection{Method 1: Wasserstein GAN}\label{IMP}
In the lower left of Fig.~\ref{fig:overview_pic}, the discriminator model of this method is shown as \textbf{discriminator1} $D_1$.
$D_1$ is a deep CNN with residual blocks, which takes a sequence of word distributions as input and outputs a score.
The discriminator loss $D_{loss}$ is
\begin{equation*} \label{eq:WGAN}
  \begin{aligned}
  D_{loss} = \frac{1}{K} \sum^{K}_{k=1} D_1(G(x^{(k)})) - \frac{1}{K} \sum^{K}_{k=1} D_1(y^{real(k)}) \\  
  + \beta_{1} \frac{1}{K}\sum_{k=1}^{K} (\Delta _{y^{i(k)}} D_1(y^{i(k)}) - 1)^{2}, 
  \end{aligned}
\end{equation*}
where $K$ denotes the number of training examples in a batch, and $k$ denotes the $k$-th example. 
The last term is the gradient penalty~\citep{improved}.
We interpolate the generator output layer $G(x)$ and the real sample $y^{real}$, and apply the gradient penalty to the interpolated sequence $y^{i}$. 
$\beta_{1}$ determines the gradient penalty scale.
In Equation~(\ref{eq:Gloss}), for WGAN, the generator maximizes $D_{loss}^{\prime}$:
\begin{equation}
D_{loss}^{\prime} =  \frac{1}{K}\sum^{K}_{k=1}D_1(G(x^{(k)})).
\end{equation}
\subsection{Method 2: Self-Critic Adversarial REINFORCE}\label{ARL}
In this section, we describe in detail the proposed adversarial REINFORCE method. The core idea is we use the LSTM discriminator to evaluate the current quality of the generated sequence 
$\{y^{s}_{1},y^{s}_{2},...,y^{s}_{i}\}$ at each time step $i$.
The generator knows that compared to the last time step, as the generated sentence either improves or worsens, it can easily find the problematic generation step in a long sequence, and thus fix the problem easily.\vspace{-1.0mm}

\subsubsection{Discriminator 2}
\begin{figure}[ht]
  \centering
    \includegraphics[width=0.9\linewidth]{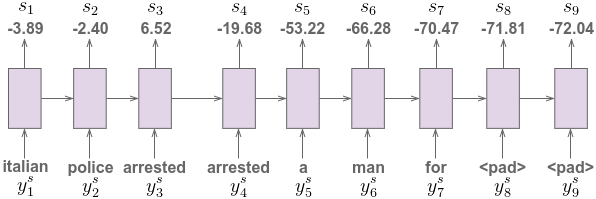}
  \caption{When the second \emph{arrested} appears, as the sentence becomes ungrammatical, the discriminator determines that this example comes from the generator. Hence, after this time-step, it outputs low scores.}
  \label{fig:example_dis}
\end{figure}

\vspace{-1.0mm}
\label{subsec:dis2}
As shown in Fig.~\ref{fig:overview_pic}, the  \textbf{discriminator2} $D_2$  is a  unidirectional LSTM network  which takes a discrete word sequence as input. At time step $i$, given input word $y^{s}_{i}$ it predicts the current score $s_{i}$ based on the sequence $\{y_{1},y_{2},...,y_{i}\}$. The score is viewed as the quality of the current sequence. 
An example of discriminator regularized by weight clipping\citep{wgan} is shown in Fig.~\ref{fig:example_dis}.

\setlength{\belowcaptionskip}{-10pt}

In order to compute the discriminator loss $D_{loss}$, we sum the scores $\{s_{1},s_{2},...,s_{N}\}$ of the whole sequence $y^{s}$ to yield
\setlength{\abovedisplayskip}{3pt}
\setlength{\belowdisplayskip}{3pt}
\begin{gather*}
    D_2(y^{s}) = \frac{1}{N} \sum_{n=1}^{N} s_{n}.
\end{gather*}
where N denotes the generated sequence length. 
Then,  the loss of discriminator is
\setlength{\abovedisplayskip}{3pt}
\setlength{\belowdisplayskip}{3pt}
\begin{gather*}
    D_{loss} =  \frac{1}{K}\sum_{k=1}^{K} D_2(y^{s(k)}) -  \frac{1}{K}\sum_{k=1}^{K} D_2(y^{real(k)}) \\
    + \beta_{2} \frac{1}{K}\sum_{k=1}^{K}(\Delta _{y^{i(k)}} D_2(y^{i(k)}) - 1)^{2},　
\end{gather*}
Similar to previous section, the last term is gradient penalty term.
With the loss mentioned above, the discriminator attempts to quickly determine whether the current sequence is real or fake. 
The earlier the timestep discriminator determines whether the current sequence is real or fake, the lower its loss. 

\begin{table*}[h]
\centering
\begin{tabular}{|c|l|c|c|c|c|}
\hline
Task & Labeled & Methods &R-1 & R-2 &R-L\\ 
\hline\hline
\multirow{4}{*}{(A)Supervised} & \multirow{4}{*}{3.8M} & (A-1)Supervised training on generator& $33.19$ & $14.21$ & $30.50$\\ \cline{3-6}
{} & {} & (A-2) \citep{Rush-attn}\ding{61} & $29.76$ & $11.88$ & $26.96$ \\ \cline{3-6}
{} & {} & (A-3) \citep{RAS}\ding{61} & $33.78$ & $15.97$ & $31.15$ \\ \cline{3-6}
{} & {} & (A-4) \citep{selective}\ding{61} & \textbf{36.15} & \textbf{17.54} & \textbf{33.63}\\ \cline{1-6}
(B) Trivial baseline & $0$ & (B-1) Lead-8 &  $21.86$ & $7.66$ & $20.45$\\ \cline{1-6}
\multirow{3}{*}{(C) Unpaired} & \multirow{3}{*}{$0$} & (C-1) Pre-trained generator & $21.26$ & $5.60$ & $18.89$\\ \cline{3-6}
{} & {} & (C-2) WGAN & $28.09$ & $9.88$ & $25.06$ \\ \cline{3-6}
{} & {} & (C-3) \textbf{Adversarial REINFORCE} & \textbf{28.11} & \textbf{9.97} & \textbf{25.41}\\ \cline{1-6}
\multirow{8}{*}{(D) Semi-supervised} & \multirow{2}{*}{10K} & (D-1) WGAN &  $29.17$ & $10.54$ & $26.72$\\ \cline{3-6}
{} & {} & (D-2) \textbf{Adversarial REINFORCE} & \textbf{30.01} & \textbf{11.57} & \textbf{27.61}\\ \cline{2-6}
{} & \multirow{3}{*}{500K} & (D-3)\citep{language}\ding{61} & $30.14$ & $12.05$ & $27.99$\\ \cline{3-6}
{} & {} & (D-4) WGAN & $32.50$ & $13.65$ & $29.67$\\ \cline{3-6}
{} & {} & (D-5) \textbf{Adversarial REINFORCE} & \textbf{33.33} & \textbf{14.18} & \textbf{30.48}\\ \cline{2-6}
{} & \multirow{3}{*}{1M} & (D-6)\citep{language}\ding{61} &  $31.09$ & $12.79$ & $28.97$\\ \cline{3-6}
{} & {} & (D-7) WGAN & $33.18$ & $14.19$ & $30.69$\\ \cline{3-6}
{} & {} & (D-8) \textbf{Adversarial REINFORCE} & $\textbf{34.21}$ & $\textbf{15.16}$ & $\textbf{31.64}$\\ \cline{2-6}
\cline{1-6}
 & \multirow{3}{*}{0} &(E-1) Pre-trained generator& $21.49$ & $6.28$ & $19.34$\\ \cline{3-6}
\multirow{1}{*}{(E) Transfer learning} & {} &(E-2) \textbf{WGAN} & $25.11$ & $7.94$ & $23.05$ \\ \cline{3-6}
{} & {} & (E-3) \textbf{Adversarial REINFORCE} & \textbf{27.15} & \textbf{9.09} & \textbf{24.11}\\ 
\hline
\end{tabular}
\vspace{-5pt}
\caption[The caption]{Average F1 ROUGE scores on English Gigaword. 
R-1, R-2 and R-L refers to ROUGE 1, ROUGE 2 and ROUGE L respectively. 
Results marked with \ding{61} are obtained from corresponding papers. In part (A), the model was trained supervisedly. In row (B-1), we select the article's first eight words as its summary. 
Part (C) are the results obtained without paired data.
In part (D), we trained our model with few labeled data. 
In part (E), we pre-trained generator on CNN/Diary and used the summaries from CNN/Diary as real data for the discriminator.}
\label{table:english_giga}
\end{table*}

\subsubsection{Self-Critical Generator}\label{sc}\vspace{-2.0mm}
Since we feed a discrete sequence $y^{s}$ to the discriminator, the gradient from the discriminator cannot directly back-propagate to the generator. Here, we use the policy gradient method. At timestep $i$, we use the $i-1$ timestep score $s_{i-1}$ from the discriminator as its self-critical baseline. The reward  $r^{D}_{i}$ evaluates whether the quality of sequence in timestep $i$ is better or worse than that in timestep $i-1$. The generator reward $r^{D}_{i}$ from $D_2$ is\vspace{-1.0mm}
\begin{equation*}　%
  r^{D}_{i}=\left\{
  \begin{aligned}
      &s_{i} \quad &\textrm{if i = 1}\\
      &s_{i} - s_{i-1} \quad &\textrm{otherwise.}\\ 
  \end{aligned}
  \right.
\end{equation*}
However, some sentences may be judged as bad sentences at the previous timestep, but at later timesteps judged as good sentences, and vice versa. Hence we use the discounted expected reward $d$ with discount factor $\gamma$ to calculate the discounted reward $d_{i}$ at time step $i$ as\vspace{-1.0mm} 
\begin{gather*}
  d_{i}=\sum_{j=i}^{N} \gamma^{j-i} r^{D}_{j}.
\end{gather*}
To maximize the expected discounted reward $d_{i}$, the loss of generator is:
\begin{equation}
G^{\prime}_{loss} = -E_{y^{s}_{i} \sim p_{G}(y^{s}_{i}|y^{s}_{1},...,y^{s}_{i-1},x)}[d_{i}]. \label{eq:D2primeloss}
\end{equation}
We use the likelihood ratio trick to approximate the gradient to minimize (\ref{eq:D2primeloss}).\vspace{-2.0mm}

\section{Experiment}\vspace{-2.0mm}

Our model was evaluated on the English/Chinese Gigaword datasets and CNN/Daily Mail dataset.
In Section \ref{exp:en},\ref{exp:semi} and \ref{exp:transfer}, the experiments were conducted on English Gigaword, while  the experiments were conducted on CNN/Daily Mail dataset and Chinese Gigaword dataset respectively in Sections ~\ref{exp:CNN} and~\ref{exp:chinese}.
We used ROUGE\citep{ROUGE} as our evaluation metric.\footnote{We used pyrouge package with option -m -n 2 -w 1.2 to compute ROUGE score for all experiments.}
During testing, when using the generator to generate summaries, we used beam search with beam size=5, and we eliminated repetition.
We provide the details of the  implementation and corpus re-processing respectively in Appendix~\ref{app:implementation} and~\ref{app:corpus}.

Before jointly training the whole model, we pre-trained the three major components~-- generator, discriminator, and reconstructor~-- separately. 
First, we pre-trained the generator in an unsupervised manner so that the generator would be able to somewhat grasp the semantic meaning of the source text.
The details of the pre-training are in Appendix~\ref{app:pre}.
We pre-trained the discriminator and reconstructor respectively with the pre-trained generator's output to ensure that these two critic networks provide good feedback to the generator. \vspace{-2.0mm}

\subsection{English Gigaword}\vspace{-2.0mm}\label{exp:en}
The English Gigaword is a sentence summarization dataset which contains the first sentence of each article and its corresponding headlines.
The preprocessed corpus contains 3.8M training pairs and 400K validation pairs. 
We trained our model on part of or fully unparalleled data on 3.8M training set. 
To have fair comparison with previous works, the following experiments were evaluated on the 2K testing set same as \citep{Rush-attn,language}. 
We used the sentences in article headlines as real data for discriminator\footnote{Instead of using general sentences as real data for discriminator, we chose sentences from headlines because they have their own unique distribution.}. 
As shown in the following experiments, the headlines can even come from another set of documents not related to the training documents.

The results on English Gigaword are shown in Table~\ref{table:english_giga}.
WGAN and adversarial REINFORCE refer to the adversarial training methods  mentioned in Sections~\ref{IMP} and~\ref{ARL} respectively.
Results trained by full labeled data are in part (A). 
In row (A-1), We trained our generator by supervised training.
Compared with the previous work~\citep{selective}, we used simpler model  and smaller vocabulary size.
We did not try to achieve the state-of-the-art results because the focus of this work is unsupervised learning, and the proposed approach is independent to the summarization models used.
In row (B-1), we simply took the first eight words in a document as its summary.

The results for the pre-trained generator with method mentioned in Appendix.\ref{app:pre} is shown in row (C-1).
In part (C), we directly took the sentences in the summaries of Gigaword as the training data of discriminator.
Compared with the pre-trained generator and the trivial baseline , the proposed approach (rows (C-2) and (C-3)) showed good improvement.
In Fig.~\ref{fig:example1}, we provide a real example.
More examples can be found in the Appendix.\ref{app:example_giga}. \vspace{-2.0mm}

\subsection{Semi-Supervised Learning}\vspace{-2.0mm}\label{exp:semi}
In semi-supervised training, generator was pre-trained with few available labeled data. During training, we conducted teacher-forcing with labeled data on generator after several updates without labeled data. With 10K, 500K and 1M labeled data, the teacher-forcing was conducted every 25, 5 and 3 updates without paired data, respectively.
In teacher-forcing, given source text as input, the generator was teacher-forced to predict the human-written summary of source text. 
Teacher-forcing can be regarded as regularization of unpaired training that prevents generator from producing unreasonable summaries of source text. 
We found that if we teacher-forced generator too frequently, generator would overfit on training data since we only used very few labeled data on semi-supervised training.\vspace{-0.5mm}

The performance of semi-supervised model in English Gigaword regarding available labeled data is shown in Table~\ref{table:english_giga} part (D).
We compared our results with~\citep{language} which was the previous state-of-the-art method on semi-supervised summarization task under the same amount of labeled data.
With both 500K and 1M labeled data, our method performed better.
Furthermore, with only 1M labeled data, using adversarial REINFORCE even outperformed supervised training in Table~\ref{table:english_giga}~(A-1) with the whole 3.8M labeled data.

\setlength{\belowcaptionskip}{-5pt}
\begin{figure}[ht]
  \centering
    \includegraphics[width=1.0\linewidth]{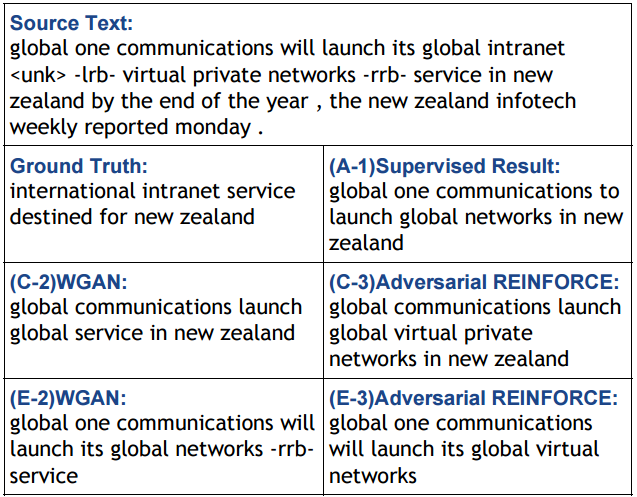}
  \vspace{-20pt}
  \caption{Real examples with methods referred in Table~\ref{table:english_giga}. The proposed methods generated summaries that grasped the core idea of the articles.}
  \label{fig:example1}
\end{figure}

\begin{table*}[h]
\centering
\begin{tabular}{|l|l|c|c|c|}
\hline
\multicolumn{2}{|c|}{Methods} & R-1 & R-2 &R-L\\ 
\hline\hline
\multirow{2}{*}{(A)Supervised} & (A-1)Supervised training on our generator& $38.89$ & $13.74$ & $29.42$\\ \cline{2-5}
{} & (A-2) \citep{getto}\ding{61} & $39.53$ & $17.28$ & $36.38$ \\ \cline{1-5}
\multicolumn{2}{|l|}{(B)Lead-3 baseline \citep{getto}\ding{61}}  &$40.34$ & $17.70$ & $36.57$\\ \cline{1-5}
\multirow{3}{*}{(C) Unpaired} & (C-1) Pre-trained generator& $29.86$ & $5.14$ & $14.66$\\ \cline{2-5}
{} & (C-2)  WGAN & $35.14$ & \textbf{9.43} & \textbf{21.04}\\ \cline{2-5}
{} & (C-3)  \textbf{Adversarial REINFORCE} & \textbf{35.51} & $9.38$ & $20.98$\\ \cline{2-5}
\hline
\end{tabular}
\vspace{-5pt}
\caption{F1 ROUGE scores on CNN/Diary Mail dataset. 
In row (B), the first three sentences were taken as summaries.
Part (C) are the results obtained without paired data. The results with symbol \ding{61} are directly obtained from corresponding papers.}
\label{table:CNN}
\end{table*}

\subsection{CNN/Daily Mail dataset} \label{exp:CNN}
The CNN/Daily Mail dataset is a long text summarization dataset which is composed of news articles paired with summaries.
We evaluated our model on this dataset because it's a popular benchmark dataset, and we want to know whether the proposed model works on long input and long output sequences.
The details of corpus pre-processing can be found in Appendix.\ref{app:corpus} .
In unpaired training, to prevent the model from directly matching the input articles to its corresponding summaries, we split the training pairs into two equal sets, one set only supplied articles and the other set only supplied summaries. 

The results are shown in Table~\ref{table:CNN}.
For supervised approaches in part (A), although our seq2seq model was similar to \citep{getto}, due to the smaller vocabulary size (we didn't tackle out-of-vocabulary words), simpler model architecture, shorter output length of generated summaries, there was a performance gap between our model and the scores reported in \citep{getto}. 
Compared to the lead-3 baseline in  part (B) which took the first three sentences of articles as summaries, the seq2seq models fell behind. 
That was because news writers often put the most important information in the first few sentences, and thus even the best abstractive summarization model only slightly beat the lead-3 baseline on ROUGE scores. However, during pre-training or training we didn't make assumption that the most important sentences are in first few sentences.

We observed that our unpaired model yielded decent ROUGE-1 score, but it yielded lower ROUGE-2 and ROUGE-L score. 
That was probably because the length of our generated sequence was shorter than ground truth, and our vocabulary size was small. 
Another reason was that the generator was good at selecting the most important words from the articles, but sometimes failed to combine them into reasonable sentences because it's still difficult for GAN to generate long sequence.
In addition, since the reconstructor only evaluated the reconstruction loss of whole sequence, as the generated sequence became long, the reconstruction reward for generator became extremely sparse. 
However, compared to pre-trained generator (rows (C-2), (C-3) v.s. (C-1)), our model still enhanced the ROUGE score. 
An real example of generated summary can be found at Appendix.\ref{app:example_giga} Fig.\ref{fig:CNN_example1} .

\begin{table*}[h]
\centering
\begin{tabular}{|l|l|c|c|c|}
\hline
\multicolumn{2}{|c|}{Methods} & R-1 & R-2 &R-L\\ 
\hline\hline
\multicolumn{2}{|l|}{(A) Training with paired data (supervised) }  & $49.62$ & $34.10$ & $46.42$\\ \cline{1-5}
\multicolumn{2}{|l|}{(B)Lead-15 baseline}  &30.08&18.24&27.74\\ \cline{1-5}
\multirow{3}{*}{(C) Unpaired} & (C-1) Pre-trained generator& $28.36$ & $16.73$ & $26.48$\\ \cline{2-5}
{} & (C-2)  WGAN & $38.15$ & $24.60$ & $35.27$\\ \cline{2-5}
{} & (C-3)  \textbf{Adversarial REINFORCE} & \textbf{41.25} & \textbf{26.54} & \textbf{37.76}\\ \cline{2-5}
\hline
\end{tabular}
\vspace{-5pt}
\caption{F1 ROUGE scores on Chinese Gigaword. 
In row (B), we selected the article's first fifteen words as its summary.
Part (C) are the results obtained without paired data.}
\label{table:chinese_giga}
\end{table*}

\subsection{Transfer Learning} \label{exp:transfer} \vspace{-2.0mm}
The experiments conducted up to this point required  headlines unpaired to the documents but in the same domain to train discriminator.
In this subsection, we generated the summaries from English Gigaword (target domain), but  the summaries for discriminator were from  CNN/Daily Mail dataset (source domain).

The results of transfer learning are shown in Table.~\ref{table:english_giga} part (E). 
Table~\ref{table:english_giga}~(E-1) is the result of pre-trained generator and the poor pre-training result indicates that the data distributions of two datasets are quite different.
We find that using sentences from another dataset yields lower ROUGE scores on the target testing set (parts (E) v.s. (C)) due to the mismatch word distributions between the summaries of the source and target domains.
However, the discriminator still regularizes the  generated word sequence. 
After unpaired training, the model enhanced the ROUGE scores of the pre-trained model (rows (E-2), (E-3) v.s. (E-1)) and it also surpassed the trivial baselines in part (B).

\subsection{GAN Training}\vspace{-2.5mm}
In this section, we discuss the performance of two GAN training methods.
As shown in the Table~\ref{table:english_giga}, in English Gigaword, our proposed adversarial REINFORCE method performed better than WGAN. 
However, in Table~\ref{table:CNN}, our proposed method slightly outperformed by WGAN. 
In addition, we find that when training with WGAN, convergence is faster.
Because WGAN directly evaluates the distance between the  continuous distribution from generator and the discrete distribution from real data, the distribution was sharpened at an early stage in training. 
This caused generator to converge to a relatively poor place. 
On the other hand, when training with REINFORCE, generator keeps seeking the network parameters that can better fool discriminator. 
We believe that training GAN on language generation with this method is worth exploring.\vspace{-3.0mm}

\subsection{Chinese Gigaword}\vspace{-2.0mm} \label{exp:chinese}
The Chinese Gigaword is a long text summarization dataset  composed of   paired headlines and news. Unlike the input news in English Gigaword, the news in Chinese Gigaword consists of several sentences.
The results are shown in Table~\ref{table:chinese_giga}.\vspace{-0.5mm}
Row (A) lists the results using 1.1M document-summary pairs to directly train the generator without the reconstructor and discriminator: this is the upper bound of the proposed approach.
In row (B), we simply took the first fifteen words in a document as its summary. The number of words was chosen to optimize the evaluation metrics.
Part (C) are the results obtained in the scenario without paired data.
The discriminator took the summaries in the training set as real data. 
We show the results of the pre-trained generator in row (C-1); rows (C-2) and (C-3) are the results for the two GAN training methods respectively.
We find that despite the performance gap between the unpaired and supervised methods (rows (C-2), (C-3) v.s. (A)), the proposed method yielded much better performance than the trivial baselines (rows (C-2), (C-3) v.s. (B)). \vspace{-2.5mm}

\section{Conclusion and Future Work}\vspace{-2.5mm}
Using GAN, we propose a model that encodes text as a human-readable summary, learned without document-summary pairs. 
In future work, we hope to use extra discriminators to control the style and sentiment of the generated summaries. 

\bibliography{emnlp2018}
\bibliographystyle{acl_natbib_nourl}

\clearpage
\appendix
\section{Implementation}
\label{app:implementation}
\textbf{Network Architecture.} The model architecture of generator and reconstructor is almost the same except the length of input and output sequence. We adapt model architecture for our generator and reconstructor from ~\citep{getto} who used hybrid-pointer network for text summarization.
The hybrid-pointer networks of generator and reconstructor are all composed of two one-layer unidirectional LSTMs as its encoder and decoder, respectively, with a hidden layer size of 600.
Since we use two kinds of methods on adversarial training, there are two discriminators with different model architecture. In the Section~\ref{IMP}, the discriminator is composed of four residual blocks with 512 hidden dimensions. While in Section~\ref{ARL}, we use only one layer unidirectional LSTM with a hidden size of 512 as our discriminator.

\noindent\textbf{Details of Training.} In all experiments except in Section \ref{exp:transfer} , we set the weight $\alpha$ in (\ref{eq:Gloss}) controlling $R_{loss}$  to 25. In Section \ref{exp:transfer}, to prevent generator from overfitting to sentences from CNN/Daily Mail summary, we set the weight $\alpha$ to 50 which was larger than other experiments. 
We find that the if the value of $\alpha$ is too large, generator will start to generate output unlike human-written sentences. 
On the other hand, if the value of $\alpha$ is too small, the sentences generated by generator will sometimes become unrelated to input text of generator. 
For all the experiments, the baseline $b$ in (\ref{eq:R_reward}) gradually decreases from $0.25$ to zero within 10000 updates on generator. 

We set the weight $\beta_{1}$ of the gradient penalty in Section~\ref{IMP} to 10, and used RMSPropOptimizer with a learning rate of 0.00001 and 0.001 on the generator and discriminator, respectively. 
In Section~\ref{subsec:dis2}, the weight $\beta_{2}$ of gradient penalty terms was $1.0$, and used RMSPropOptimizer with a learning rate of 0.00001 and 0.001 on the generator and discriminator, respectively. 
It's also feasible to apply weight clipping in discriminator training, but the performance of gradient penalty trick was better. 

\section{Corpus Pre-processing} \label{app:corpus}
\begin{itemize}
\item English Gigaword: We used the script of \citep{Rush-attn} to construct our training and testing datasets.
The vocabulary size was set to 15K in all experiments.
\item CNN/Diary Mail: 
We obtained 287227 training pairs, 13368 validation pairs and 11490 testing pairs identical to \citep{getto} by using the scripts provided by \citep{getto}.  To make our model easier to train, during training and testing time, we truncated input articles to 250 tokens (original articles has 781 tokens on average) and restricted the length of generator output summaries (original summaries has 56 tokens on average) to 50 tokens. The vocabulary size was set to $15k$.
\item Chinese Gigaword: The Chinese Gigaword is a long text summarization dataset which is composed of 2.2M paired data of headlines and news. We preprocessed the raw data as following. First, we selected the 4K most frequent Chinese characters to form our vocabulary. We filtered out headline-news pairs with excessively long or short news segments, or that contained too many out-of-vocabulary Chinese characters, yielding 1.1M headline-news pairs from which we randomly selected 5K headline-news pairs as our testing set, 5K headline-news pairs as our validation set, and the remaining pairs as our training set. During training and testing, the generator took the first 80 Chinese characters of the source text as input. 
\end{itemize}

\section{Model Pre-training} \label{app:pre}
As we found that the different pre-training methods for the generator influenced final performance dramatically in all of the experiments, we felt it was important to find a proper unsupervised pre-training method to help the machine grasp semantic meaning. 
The summarization tasks on two datasets is different: One is sentence summarization, while the other is long text summarization. 
Therefore, we used the different pre-training strategies on two datasets described below.
\begin{itemize}
\item CNN/Diary Mail:
The CNN/Diary Mail is a long text summarization dataset in which the source text consists of several sentences. Given the previous $i-1$ sentences ${sent_{0},sent_{1},...,sent_{i-1}}$ from the source text, the generator predicted the next four sentences ${sent_{i},sent_{i+1},..,sent_{i+3}}$ in the source text as its pre-training target. 
If more than 40\% of the words in target sentences ${sent_{i},sent_{i+1},...,sent_{i+3}}$ did not appear in the given text, we filtered out this pre-training sample pair.
This pre-training method allowed the generator to capture the important semantic meanings of the source text.
Although the first few sentences of articles in CNN/Diary Mail contains the main information of articles, we hope we can provide a more general pre-training method which don't have any assumption of dataset and can be easily applied to other datasets. 
\item Chinese Gigaword:
The pre-training method of Chinese Gigaword was similar to CNN/Diary Mail except that generator predicted the next sentence instead of next consecutive four sentences. 
\item English/Chinese Gigaword: 
As the source text of English Gigaword is made up of only one sentence, it is not feasible to split the last sentence from the source text; hence the previous pre-training method on Chinese Gigaword is not appropriate for this dataset. To properly initialize the set, we randomly selected 6 to 11 consecutive words in the source text, after which we randomly swapped 70\% of the words in the source text. Given text with incorrect word arrangements, the generator predicted the selected words in the correct arrangement. We pre-trained in this way because we expect the generator to initialize with a rough language model.
In Chinese Gigaword we also conducted experiments on pre-training in this manner, but the results were not as good as those shown  in the part (C) of Table~\ref{table:chinese_giga}.
In addition, we also used the retrieved paired data in row (B-1) in Table~\ref{table:english_giga} to pre-train the generator in English Gigaword. However, pre-training generator with this method doesn't yield results better than those in Table~\ref{table:english_giga}.
\item Transfer Learning:
Before unsupervised training, the generator was pre-trained with paralleled data on CNN/Daily Mail dataset.
However, the characteristics for two datasets are different.
In English Gigaword, the articles were short and the summaries consist of only one sentence, while in CNN/Daily Mail dataset, the articles were extremely long and summaries consist of several sentences.
To overcome these differences, during pre-training time, we took the first 35-45 words in each CNN/Diary Mail article as generator input, and generator randomly predicted one of the sentences of the article's summary.
In addition, we used the one sentence from CNN/Diary Mail summaries as real data to discriminator instead full summaries. 
\end{itemize}

\clearpage
\section{Examples}\vspace{-5.0mm}
\label{app:example_giga}
\setlength{\belowcaptionskip}{-10pt}
\begin{figure}[ht]
  \centering
    \includegraphics[width=1.0\linewidth]{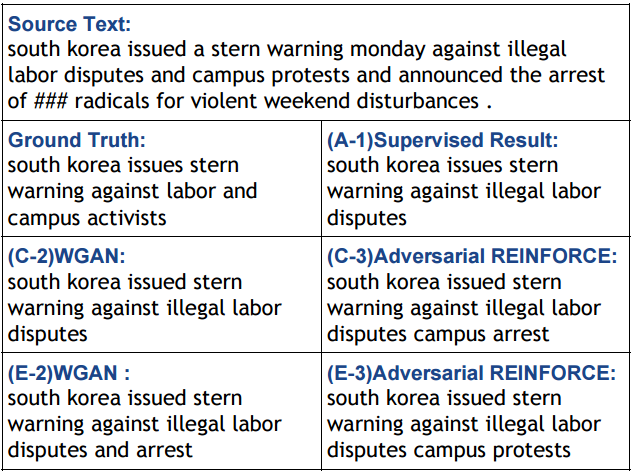}
  \caption{}
  \label{fig:example3}
\end{figure}

\setlength{\belowcaptionskip}{-20pt}
\begin{figure}[ht]
  \centering
    \includegraphics[width=1.0\linewidth]{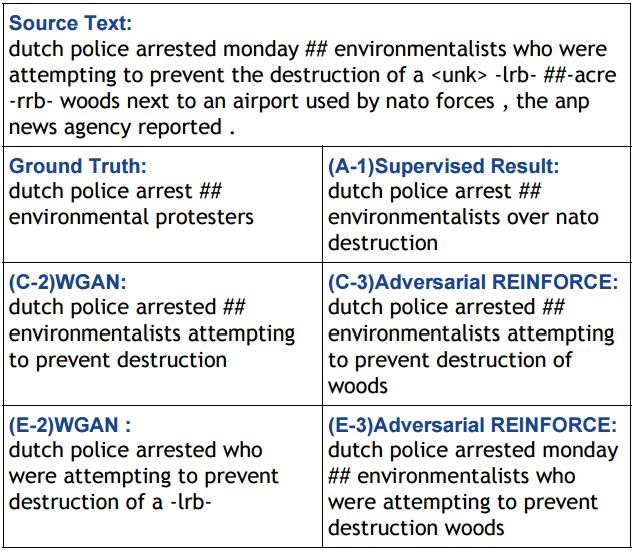}
  \caption{}
  \label{fig:example4}
\end{figure}

\setlength{\belowcaptionskip}{-1000pt}
\begin{figure}[ht]
  \centering
    \includegraphics[width=1.0\linewidth]{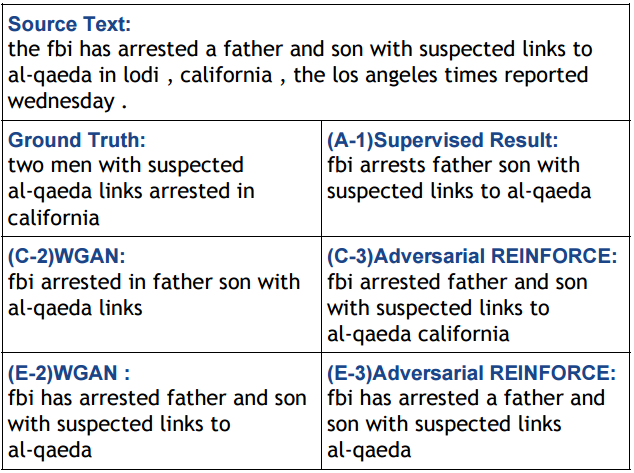}
  \caption{}
  \label{fig:example5}
\end{figure}

\setlength{\belowcaptionskip}{-20pt}
\begin{figure}[ht]
  \centering
    \includegraphics[width=1.0\linewidth]{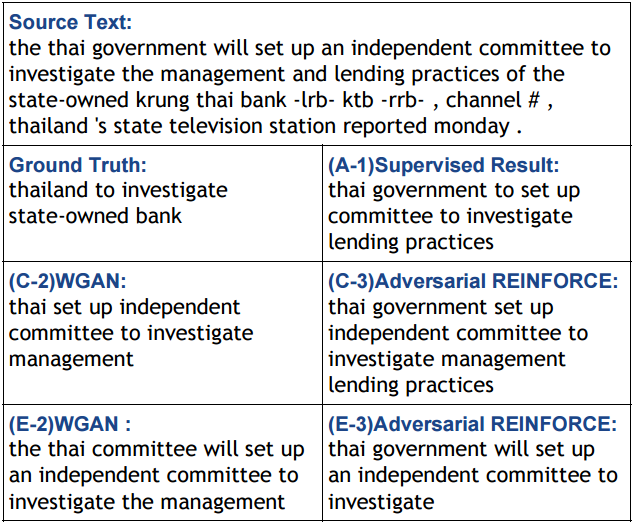}
  \caption{}
  \label{fig:example6}
\end{figure}

\setlength{\belowcaptionskip}{-2pt}
\begin{figure}[ht]
  \centering
    \includegraphics[width=1.0\linewidth]{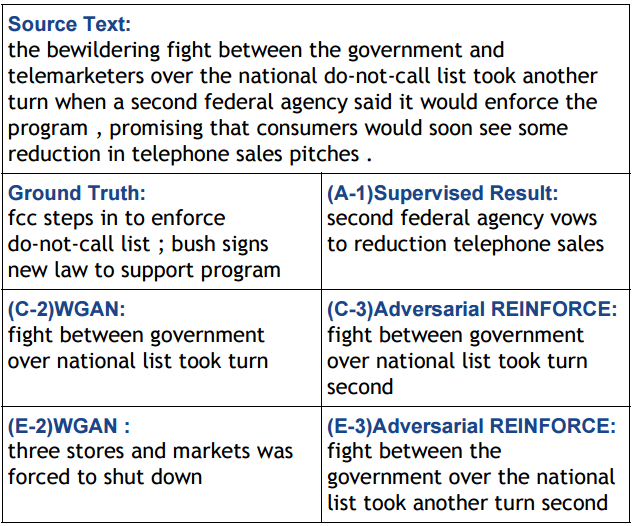}
  \caption{}
  \label{fig:example7}
\end{figure}

\begin{figure}[ht]
  \setlength{\belowcaptionskip}{-1pt}
  \centering
    \includegraphics[width=1.0\linewidth]{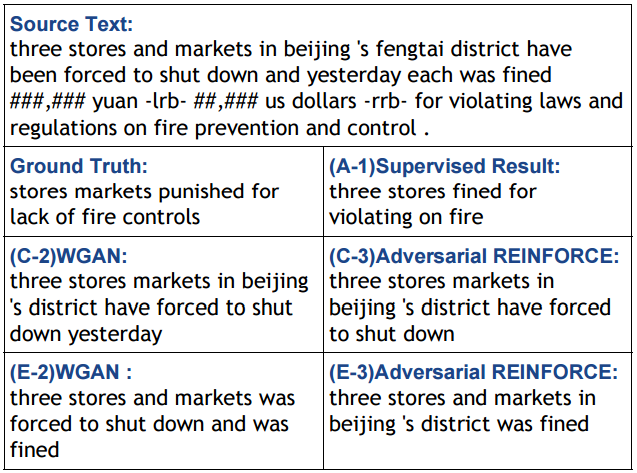}
  \caption{}
  \label{fig:example8}
\end{figure}

\begin{figure*}[ht]
  \centering
    \includegraphics[width=0.7\linewidth]{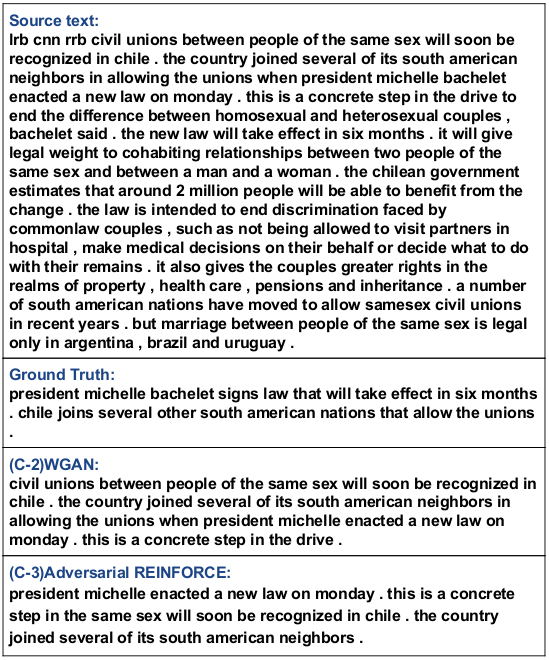}
  \caption{An example of generated summary of CNN/Diary Mail.}
  \label{fig:CNN_example1}
\end{figure*}

\end{document}